\pdfoutput=1

\documentclass[11pt]{article}

\usepackage[]{emnlp2021}

\usepackage{times}
\usepackage{latexsym}

\usepackage[T1]{fontenc}

\usepackage[utf8]{inputenc}

\usepackage{microtype}

\usepackage{amsmath}
\usepackage{amssymb}
\usepackage{amsfonts}
\usepackage{makecell}
\usepackage{graphicx}
\usepackage{subfig}
\usepackage{stfloats}  
\usepackage{booktabs}
\graphicspath{{Figure/}}
\definecolor{my_blue}{RGB}{41,50,225}

\title{Amendable Generation for Dialogue State Tracking}

\author{Xin Tian, Liankai Huang, Yingzhan Lin, Siqi Bao, Huang He,\\
  \textbf{Yunyi Yang, Hua Wu, Fan Wang, Shuqi Sun} \\
  Baidu Inc., China \\
  \texttt{\{tianxin06,huangliankai,linyingzhan01,baosiqi,hehuang,} \\
  \texttt{yangyunyi01,wu\_hua,wang.fan,sunshuqi01\}@baidu.com}
}

\begin{document}
\maketitle
\begin{abstract}
In task-oriented dialogue systems, recent dialogue state tracking methods tend to perform one-pass generation of the dialogue state based on the previous dialogue state. The mistakes of these models made at the current turn are prone to be carried over to the next turn, causing error propagation. 
In this paper, we propose a novel \textbf{A}mendable \textbf{G}eneration for \textbf{D}ialogue \textbf{S}tate \textbf{T}racking (\textbf{AG-DST}), which contains a two-pass generation process: (1) generating a primitive dialogue state based on the dialogue of the current turn and the previous dialogue state, and (2) amending the primitive dialogue state from the first pass. 
With the additional amending generation pass, our model is tasked to learn more robust dialogue state tracking by amending the errors that still exist in the primitive dialogue state, which plays the role of reviser in the double-checking process and alleviates unnecessary error propagation.
Experimental results show that AG-DST significantly outperforms previous works in two active DST datasets (MultiWOZ 2.2 and WOZ 2.0), achieving new state-of-the-art performances.
\end{abstract}

\section{Introduction}

Dialogue state tracking (DST) is a crucial task in task-oriented dialogue systems, as it affects database query results as well as the subsequent policy prediction~\citep{chen2017survey}. It extracts users' goals at each turn of the conversation and represents them in the form of a set of (\textit{slot}, \textit{value}) pairs, i.e., dialogue state.

Traditional methods of DST mainly rely on a predefined ontology which includes all possible slots and corresponding values. These models predict the value for each slot as a classification problem~\citep{mrksic-etal-2017-neural,zhong-etal-2018-global,ramadan-etal-2018-large}. However, in practical applications, some slot values appearing in the conversations cannot be predefined, and it is infeasible to acquire a fully predefined ontology or transfer to other domains with fixed predefined ontology.
To address such challenges, open-vocabulary DST has been proposed, where the value of each slot is directly generated or extracted from the dialogue history~\citep{Chao2019,NEURIPS2020_e9462095,ham-etal-2020-end,heck-etal-2020-trippy}.
Although this approach offers scalability and is capable of handling unseen slot values, many of the previous models are not efficient enough as they need to predict the dialogue state from scratch based on the dialogue history.

On the merits of utilizing the previous dialogue state as a compact representation of the previous dialogue history, some recent methods choose to take the previous dialogue state into consideration when generating the slot values. One direction is to decompose DST into two explicit sub-tasks: State Operation Prediction and Value Generation~\citep{kim-etal-2020-efficient,zeng2020multi}.
At each turn, whether or how to modify the value in the previous dialogue state is determined by the discrete operations from the state operation prediction, so the accuracy of state operation prediction holds back the overall DST performance~\citep{kim-etal-2020-efficient}. 
Another direction of recent works recasts dialogue state tracking into a single causal language model by using the dialogue of the current turn and the previous dialogue state as input sequence~\citep{lin-etal-2020-mintl,yang2021ubar}, where the current dialogue state is generated by jointly modeling the state operation prediction and value generation in a implicit fashion.
While it is more effective and reasonable to use the previous dialogue state under the Markov assumption, the mistakes of these models made during the prediction of the current turn are prone to be carried over to the next turn, causing error propagation. These carried-over mistakes are unlikely to be fixed in the next turn. Essentially, these models perform a one-pass generation process and lack a double-checking process to amend the mistakes of the current turn. Missing such amending process would result in some potential mistakes being left unfixed.



\begin{figure*}
	\centering
	\includegraphics[width=0.98\textwidth]{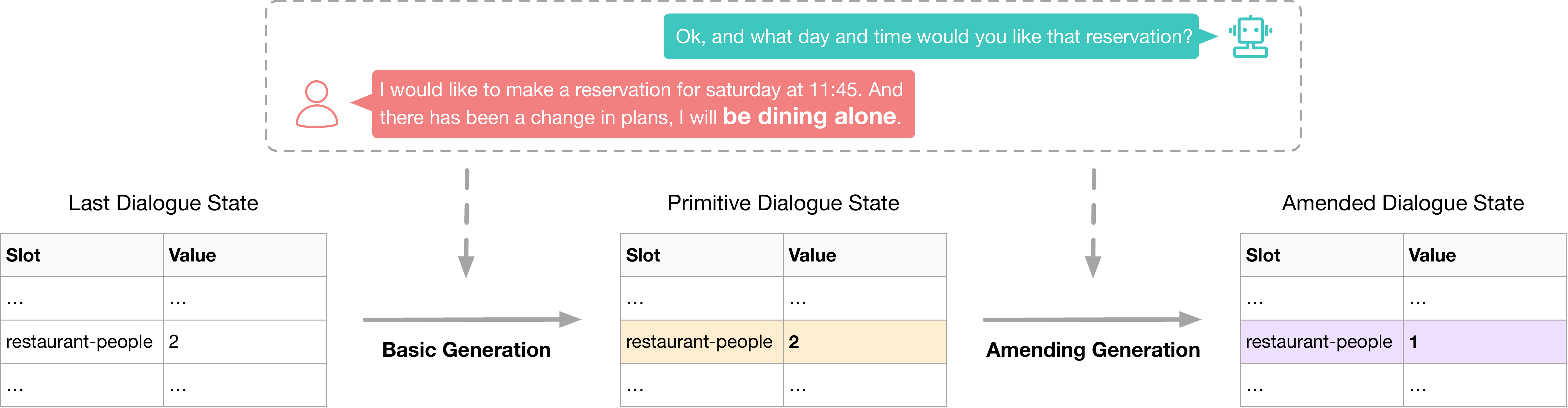}
	\caption{An example of AG-DST with two-pass generation process. In this example, the user wants to change \textit{restaurant-people} from 2 to 1 by ``\textit{... a change in plans, I will be dining alone}". In the two-pass generation process, the amending generation obtains corresponding information to correct mistake in the basic generation.}
	\label{fig:example}
\end{figure*} 

To nip the mistakes in the bud and alleviate the error propagation problem, we propose \textbf{A}mendable \textbf{G}eneration for \textbf{D}ialogue \textbf{S}tate \textbf{T}racking (\textbf{AG-DST}), a pretrained language model that generates the dialogue state based on the dialogue of the current turn and previous dialogue state. In contrast to previous one-pass generation process~\citep{kim-etal-2020-efficient,zeng2020multi,lin-etal-2020-mintl,yang2021ubar}, AG-DST employs a two-pass generation process consisting of a basic generation and an amending generation, where the first pass uses the dialogue of the current turn and the previous dialogue state to generate a primitive dialogue state, and second pass utilizes the dialogue of the current turn to amend the primitive dialogue state. 
With the additional amending generation pass, our model is tasked to learn more robust dialogue state tracking by amending the errors that still exist in the primitive dialogue state. These errors are more challenging to fix and relatively scarce during training. Therefore, we further design a negative sampling mechanism to exploit more challenging errors and facilitate the effective learning of the amending generation pass. With such two-pass generation process, AG-DST is less likely to generate false dialogue state for the next turn, and thus reduces error propagation.
Figure~\ref{fig:example} illustrates a complete dialogue state generation process of AG-DST. 

Experimental results show that AG-DST consistently outperforms all prior works on MultiWOZ 2.2 and WOZ 2.0. Especially on MultiWOZ 2.2, AG-DST achieves 57.26\% joint goal accuracy, 2.86\% higher than the previous state-of-the-art performance. Besides, we provide ablation study and the attention visualization to demonstrate the effectiveness of the amending generation, and analyze the types of mistakes that can be corrected by the amending generation. Our models and code will be released for further research.\footnote{\url{https://github.com/PaddlePaddle/Knover/tree/develop/projects/AG-DST}}

\section{Methodology}

In this section, we introduce AG-DST in the following aspects: the basic generation, the amending generation and the training objective.

\subsection{Basic Generation}

\begin{figure*}
	\centering
	\includegraphics[width=0.8\textwidth]{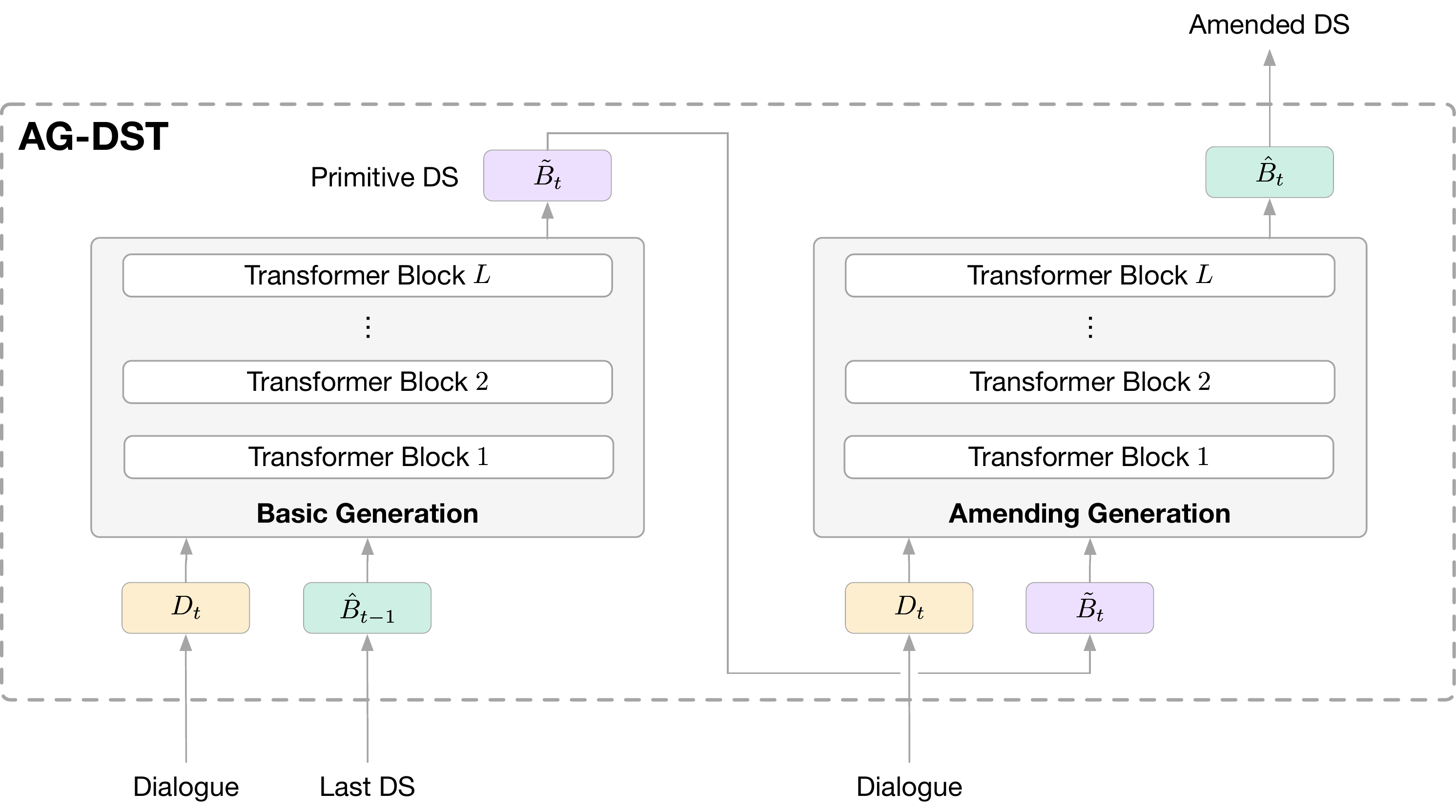}
	\caption{The overview of the proposed AG-DST. In the basic generation, AG-DST takes the dialogue of the current turn $D_t$ and the previous dialogue state $\hat{B}_{t-1}$ as input and generates the primitive dialogue state $\tilde{B}_t$. In the amending generation, AG-DST takes the dialogue of the current turn $D_t$ and the primitive dialogue state $\tilde{B}_t$ as input and outputs the amended dialogue state $\hat{B}_t$.}
	\label{fig:overview}
\end{figure*}

A dialogue with $T$ turns can be represented as $D=\{D_1,D_2,\dots,D_T\}$, where $D_t$ is the dialogue at turn $t$ consisting of system response $R_t$ and user utterance $U_t$. We denote the dialogue states at every turn as $\mathcal{B}=\{\mathcal{B}_1,\mathcal{B}_2,\dots,\mathcal{B}_T\}$. For multi-domain DST, the dialogue state at turn $t$ is denoted as $\mathcal{B}_t=\{(S^i,V_t^i)|1\leq i\leq I\}$, in which $S^i$ is the slot and $V_t^i$ is its corresponding value ($I$ is the number of all slots in different domains). Particularly, $S^i$ is represented as a special token concatenated by domain and slot (i.e. \texttt{<domain-slot>}) following most previous works. We use special tokens \texttt{<nm>} and \texttt{<dc>} to indicate \textit{not mentioned} and \textit{don't care} in slot value respectively.

We leverage the dialogue of the current turn and the previous dialogue state to generate the current dialogue state. The dialogue of the current turn $D_t$ is used in the input tokens under the Markov assumption. To a certain extent the previous dialogue state $B_{t-1}$ could be viewed as a compact representation of the previous dialogue history~\citep{kim-etal-2020-efficient}. Specifically, we denote the dialogue at turn $t$ as:
\begin{equation}
    D_t=[R_t;U_t]
\end{equation}
where $R_t$ and $U_t$ are system response and user utterance accordingly. Special tokens $\texttt{<con/>}$ and $\texttt{</con>}$ are added around $D_t$ for marking the boundary of the dialogue context, and special tokens $\texttt{<sys>}$ and $\texttt{<usr>}$ are added before $R_t$ and $U_t$ respectively to indicate the role. The dialogue state at turn $t$ is denoted as
\begin{equation}
    B_t=[B_t^1;B_t^2;\dots;B_t^I]
\end{equation}
where $B_t^i=[S^i;V_t^i]$ is the concatenation of $i$-th slot and its value. Similar to the dialogue context, two special tokens \texttt{<ds/>} and \texttt{</ds>} are added around the whole dialogue state. 

The overview of AG-DST is illustrated in Figure~\ref{fig:overview}. AG-DST is a generation model based on transformer~\citep{vaswani2017attention,li2019unified}. In the basic generation of AG-DST, the input sequence is composed of the current turn dialogue and the previous dialogue state, and the primitive dialogue state is predicted by:
\begin{equation}
    \tilde{B_t}=Transformer(D_t,B_{t-1}) \label{eq:basic_transformer}
\end{equation}
where \texttt{<gen/>} and \texttt{</gen>} are added around the whole input sequence to indicate the first pass generation process.

\begin{figure*}
	\centering
	\includegraphics[width=0.98\textwidth]{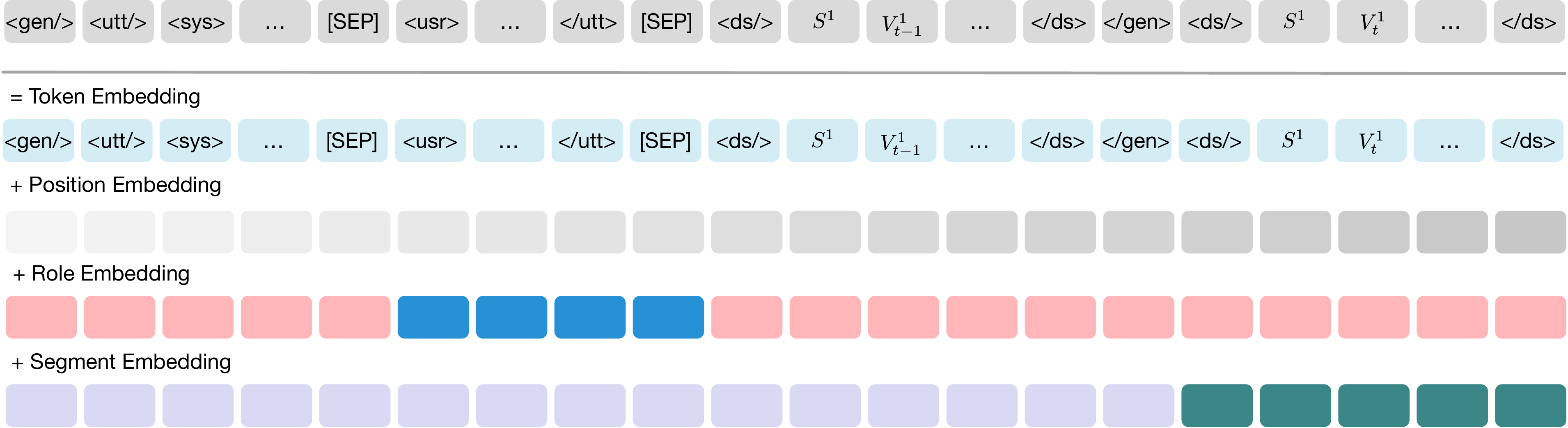}
	\caption{An example of input representation. The input embedding of each token is the sum of token embedding, position embedding, role embedding and segment embedding.}
	\label{fig:embedding}
\end{figure*}

As shown in Figure~\ref{fig:embedding}, the input embedding of each token is the sum of token embedding, position embedding, role embedding and segment embedding. Among them, position embedding is added to discriminate input token positions; role embedding is employed to distinguish the characters of the speaker in the dialogue; segment embedding is used for different types of sequence.

\subsection{Amending Generation}\label{sec:amending_generation}

To amend the potential errors in the primitive dialogue state, we propose a novel amending generation that takes the dialogue of the current turn and the primitive dialogue state predicted by the basic generation as input, and generates the amended dialogue state:
\begin{equation}
    \hat{B_t}=Transformer(D_t,\tilde{B_t}) \label{eq:amending_transformer}
\end{equation}
where the new input sequence of amending generation is consisted of the current turn dialogue and the primitive dialogue state, $\hat{B_t}$ is the amended dialogue state. The amending generation shares the same parameters with the basic generation model. To differentiate this two-pass generation process, the special tokens \texttt{<amend/>} and \texttt{</amend>} are added around the new input sequence in Equation~\ref{eq:amending_transformer} as opposed to the \texttt{<gen/>} and \texttt{</gen>} in Equation~\ref{eq:basic_transformer}.

\paragraph{Negative Sampling} 
To facilitate effective learning of the amending generation process, we propose a negative sampling strategy that actively mines the examples on which the model is prone to make mistakes (i.e., generating the wrong slot values, or failing to update some slots).
Specifically, we performs negative sampling on the dialogue state with changed slot values between turns $\Delta \mathcal{B}_t=\mathcal{B}_t-\mathcal{B}_t\cap \mathcal{B}_{t-1}$, where these slot values are randomly replaced by \texttt{<nm>}, \texttt{<dc>} or some wrong values. 
During training, the dialogue state after negative sampling is used as the primitive dialogue state $\tilde{B}_t$ in the amending generation to encourage the model to lay more emphasis on error correction.
\subsection{Training Objective}

The training objective of the basic generation is the negative log-likelihood loss given the dialogue of the current turn and the previous dialogue state:
\begin{equation}
    \mathcal{L}_{basic}=-\log P(B_t|D_t,B_{t-1})
\end{equation}
Similar to the basic generation, the loss of the amending generation is also the negative log-likelihood loss:
\begin{equation}
    \mathcal{L}_{amending}=-\log P(B_t|D_t,\tilde{B_{t}})
\end{equation}

The total objective of AG-DST is to minimize the sum of the above two losses:
\begin{equation}
    \mathcal{L}=\mathcal{L}_{basic}+\mathcal{L}_{amending}
\end{equation}

\section{Experiments}

\subsection{Datasets}

We conduct our experiments on both multi-domain dataset MultiWOZ 2.2~\citep{zang-etal-2020-multiwoz} and single-domain dataset WOZ 2.0~\citep{wen-etal-2017-network}. MultiWOZ~\citep{budzianowski-etal-2018-multiwoz} is a large-scale multi-domain dialogue dataset with human-human conversations including over 10,000 dialogues. It is a widely used benchmark for dialogue state tracking. MultiWOZ 2.2~\citep{zang-etal-2020-multiwoz} is a new version of MultiWOZ 2.1~\citep{eric2019multiwoz}, in which a number of dialogue state annotation errors across 17.3\% of the utterances have been fixed. 
Following~\citet{wu-etal-2019-transferable}, due to the absence of \textit{hospital} and \textit{police} domains in the validation and test datasets, there are only 5 domains (\textit{attraction}, \textit{hotel}, \textit{restaurant}, \textit{taxi} and \textit{train}) and 30 corresponding domain-slot pairs in our experiments. WOZ 2.0~\citep{wen-etal-2017-network} is a well-known single-domain DST dataset, where 3 slots (\textit{area}, \textit{food} and \textit{price range}) are involved in the \textit{restaurant} domain. Table~\ref{tab:datasets} summarizes the statistics of the above two datasets.

\begin{table}
\centering
\begin{tabular}{lcc}
\hline
\textbf{Statistics} & \makecell[c]{\textbf{MultiWOZ} \\ \textbf{2.2}} & \makecell[c]{\textbf{WOZ} \\ \textbf{2.0}} \\
\hline
\hline
\# domains & 5 & 1 \\
\# slots & 30 & 3 \\
\# dialogues & 10424 & 1200 \\
\# train dialogues & 8426 & 600 \\
\# valid dialogues & 998 & 200 \\
\# test dialogues & 1000 & 400 \\
Avg. turns per dialogue & 13.71 & 8.35 \\
Avg. tokens per turn & 16.86 & 13.18 \\
\hline
\end{tabular}
\caption{Statistics of the datasets in the experiments.}\label{tab:datasets}
\end{table}

\subsection{Implementation Details}\label{sec:implementation_details}

Our AG-DST approach can be easily deployed with many pre-trained generation models (such as GPT-2~\citep{radford2019language}, T5~\citep{JMLR:v21:20-074}, PLATO-2~\citep{bao-etal-2021-plato}). In this paper, we employ the GPT-2 and PLATO-2 to initialize our model parameters. GPT-2 is a large causal language model, and PLATO-2 is a large-scale open-domain dialogue model trained on the Reddit comments. Specifically, GPT-2 has 117M parameters containing 12 transformer blocks, 12 attention heads and 768 hidden units, and PLATO-2 has 310M parameters containing 24 transformer blocks, 16 attention heads and 1024 hidden units.\footnote{Unless otherwise specified in subsequent experiments, the pre-trained backbone we used is PLATO-2 as it carries out better results.} Adam optimizer~\citep{yoshua2015adam} is employed for optimization in all experiments. For the hyper-parameters we used in the best model, the PLATO-2 is fine-tuned with a dynamic batch size of 8192 tokens, a learning rate of $1e-5$, warmup steps of 1000 and a  learning rate decay rate of 0.01. For GPT-2, we set the batch size of 6, a learning rate of $5e-5$ with no warmup and learning rate decay. All the models are trained on 4 Nvidia Telsa V100 32G GPU cards for 40 epochs and early stop according to the performance on the validation set. All the reported results of AG-DST are averages over five runs.


\subsection{Baselines}

We compare our approach with the following methods:
\paragraph{Neural Belief Tracker}~\citep{mrksic-etal-2017-neural} learns the distributed representation of system responses and user utterances from pre-trained word vectors, and decides which slot-value pairs are required by the user.
\paragraph{Belief Tracking}~\cite{ramadan-etal-2018-large} proposes a model that utilizes semantic similarity between ontology terms and utterances and shares the information across domains.
\paragraph{GLAD}~\citep{zhong-etal-2018-global} uses global modules to share parameters across slots and local modules to learn the features which take into account the specific slot.
\paragraph{StateNet}~\citep{ren-etal-2018-towards} proposes a model that composes a representation of the dialogue history and measures the distances between the representation and the value vectors.
\paragraph{TRADE}~\citep{wu-etal-2019-transferable} uses a state operation with an utterance encoder and a dialogue state generator to handle the cross domain phenomenon.
\paragraph{DS-DST}~\citep{zhang-etal-2020-find} proposes a dual strategy: picklist-based and span-based. The span-based strategy includes a slot-gate classifier and span-based slot-value prediction.
\paragraph{BERT-DST}~\citep{Chao2019} adopts BERT to encode dialogue context and extracts slot values by the state operation prediction and the span prediction.
\paragraph{SOM-DST}~\citep{kim-etal-2020-efficient} proposes an efficient decoder by updating dialogue state as a memory to reduce the burden of the decoder.
\paragraph{COMER}~\citep{ren-etal-2019-scalable} adopts two sequential decoders to generate dialogue state. The first decoder is used to generate state sketch (i.e. domains and slots), and the second one is used to generate slot values by conditioning on the dialogue history and state sketch.
\paragraph{SGD-baseline}~\citep{Rastogi_Zang_Sunkara_Gupta_Khaitan_2020} adapts BERT to obtain the schema embeddings (including intents, slots and possible values of categorical slots) and utterance embeddings and uses different strategies for non-categorical and categorical slots.
\paragraph{SimpleTOD}~\citep{NEURIPS2020_e9462095} changes sub-tasks of task-oriented dialogue into a single causal language model which generates dialogue state, system action and system response successively.
\paragraph{Seq2Seq-DU}~\citep{feng-etal-2021-sequence} applies schema descriptions to deal with unseen domains with a sequence-to-sequence framework.

\subsection{Experimental Results}
We use joint goal accuracy (Joint Acc) as our main evaluation metric for dialogue state tracking. Joint goal accuracy measures the percentage of \textit{correct} in all dialogue turns, where a turn is considered as \textit{correct} if and only if all the slot values are correctly predicted. We also show the slot accuracy for each domain which measures the accuracy of all the slot values in that specific domain.

Table~\ref{tab:jga-baselines} shows the performance of the AG-DST in comparison to the baselines. We can observe that the AG-DST consistently outperforms all baselines on both MultiWOZ 2.2 and WOZ 2.0 datasets, achieving a new state-of-the-art performance. On MultiWOZ 2.2, AG-DST achieve 57.26\% joint goal accuracy, by 2.86\% significant improvement on the top of the Seq2Seq-DU~\citep{feng-etal-2021-sequence}, the latest sequence-to-sequence generation model. In addition, the domain-specific results of our approach are also provided in Table~\ref{tab:jga-domain}. 
On the single-domain dataset, WOZ 2.0, we obtain joint goal accuracy of 91.37\%, which indicates that our model is also effective in a relatively simple scenario. We will analyse the strengths of AG-DST in the subsequent section.

\begin{table}
\centering
\begin{tabular}{lcc}
\hline
\textbf{Model} & \makecell[c]{\textbf{MultiWOZ} \\ \textbf{2.2}} & \makecell[c]{\textbf{WOZ} \\ \textbf{2.0}} \\
\hline
\hline
Neural Belief Tracker & - & 84.20 \\
Belief Tracking & - & 85.50 \\
GLAD & - & 88.10 \\
StateNet & - & 88.90 \\
TRADE & 45.40$^\dagger$ & - \\
DS-DST & 51.70$^\dagger$ & - \\
BERT-DST & - & 87.70 \\
SOM-DST & 53.81$^\ddagger$ & - \\
COMER & - & 88.60 \\
SGD-baseline & 42.00$^\dagger$ & - \\
SimpleTOD & 54.02$^\ddagger$ & - \\
Seq2Seq-DU & 54.40 & 91.20 \\
\hline
AG-DST & \textbf{57.26} & \textbf{91.37} \\
\hline
\end{tabular}
\caption{Joint goal accuracy of AG-DST and baselines on MultiWOZ 2.2 and WOZ 2.0 datasets. AG-DST consistently outperforms all baselines. $^\dagger$: the results borrowed from~\citet{zang-etal-2020-multiwoz}. $^\ddagger$: our reproduction results by official code.}\label{tab:jga-baselines}
\end{table}

\section{Analysis}

\begin{table}
\centering
\begin{tabular}{lcc}
\hline
\textbf{Domain} & \textbf{Joint Acc} & \textbf{Slot Acc} \\
\hline
\hline
multi-domain & 57.26 & 97.48 \\
\hline
attraction & 89.91 & 96.26 \\
hotel & 82.31 & 97.16 \\
restaurant & 89.00 & 97.97 \\
taxi & 96.09 & 98.31 \\
train & 89.15 & 97.51 \\
\hline
\end{tabular}
\caption{Joint goal accuracy and slot accuracy of AG-DST on MultiWOZ 2.2 by domains.}\label{tab:jga-domain}
\end{table}

\subsection{Amending Generation}
As shown in Table~\ref{tab:ablation-amend}, we conduct ablation study to investigate the effectiveness of the proposed amending generation. The results show that the amending generation has a significant improvement on the basic generation, where the model gets extra joint goal accuracy of 0.87\%. When we perform negative sampling to facilitate effective learning of the amending generation process, it will lead to an increase by 1.06\%. Furthermore, we implement heuristic negative sampling which exchanges some correlated slot values (such as \textit{leave at} and \textit{arrive by}, \textit{departure} and \textit{destination}, \textit{area} in different domains) on MultiWOZ 2.2. While using this further heuristic negative sampling, the model achieves the best joint goal accuracy of 57.35\%. 
Essentially, the amending generation process amends the challenging mistakes that still exists after the basic generation and provides a more accurate prediction of the dialogue state of the current turn, which effectively alleviate the error propagation problem.
To the best of our knowledge, AG-DST is the first DST method that involves a two-pass generation process to amend the primitive dialogue state. 

\begin{figure}
	\centering
	\subfloat[Basic Generation\label{fig:basic_generation}]{\includegraphics[width=0.22\textwidth]{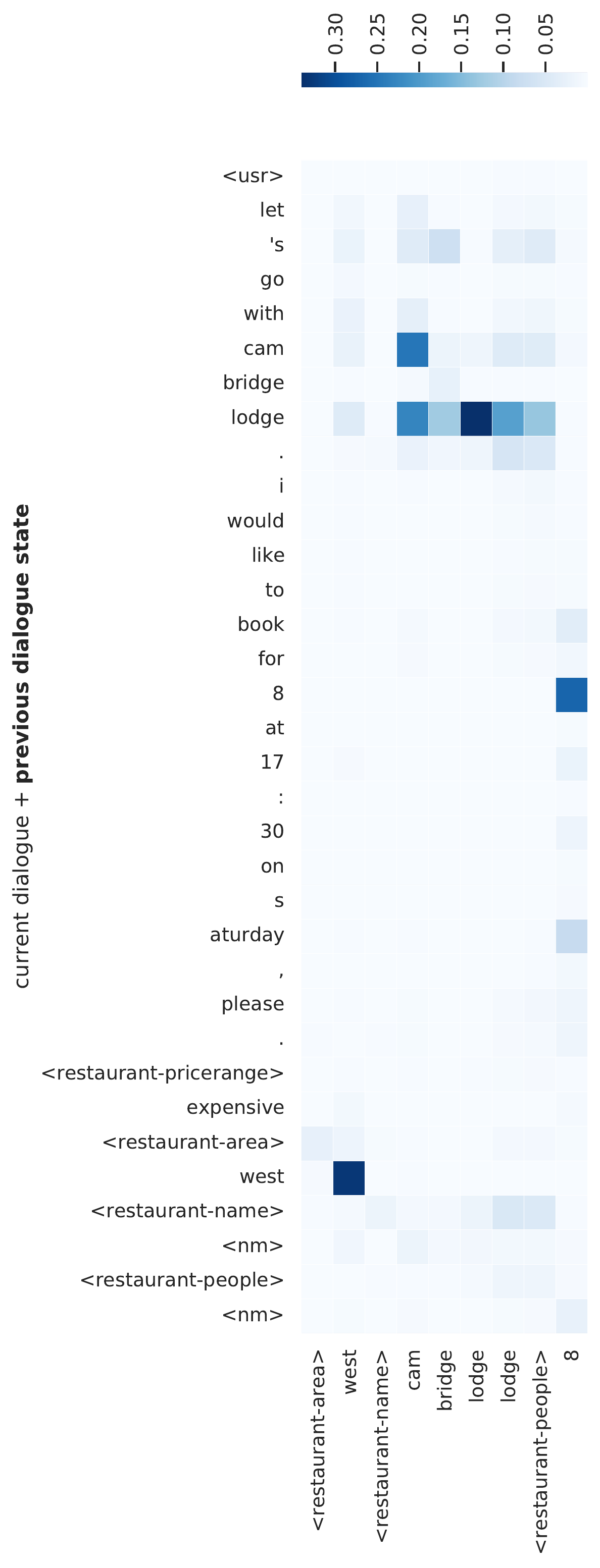}}\hspace{1 em}
	\subfloat[Amending Generation\label{fig:amending_generation}]{\includegraphics[width=0.22\textwidth]{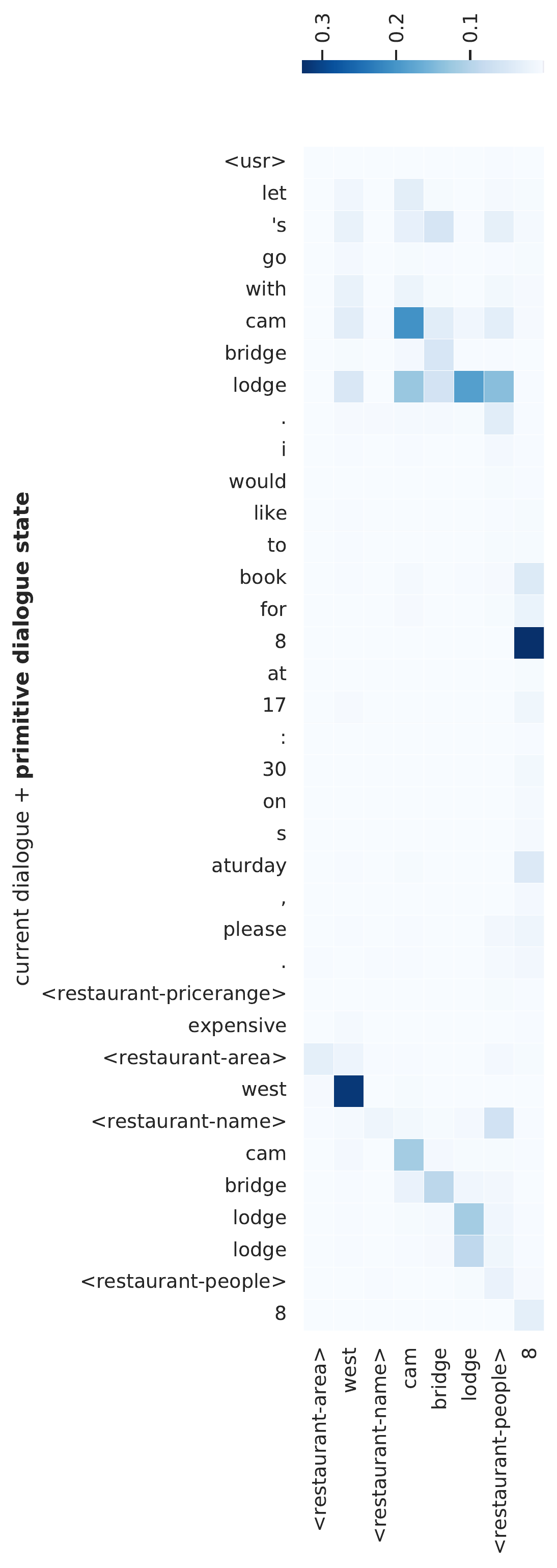}}
	\caption{The attention visualization of (a) basic generation and (b) amending generation on an example from MultiWOZ 2.2. Note that in both figures, the abscissa and ordinate are a fraction of whole input and output sequence respectively because of limited space.}
	\label{fig:attention}
\end{figure}

\begin{table}
\centering
\small
\begin{tabular}{ll}
\hline
\textbf{Model} & \textbf{Joint Acc} \\
\hline
\hline
basic generation & 56.20 \\
\hline
\quad + amending generation & 57.07 (+0.87) \\
\quad + amending generation w/ NS & 57.26 (+1.06) \\
\quad + amending generation w/ NS$^+$ & 57.35 (+1.15) \\
\hline
\end{tabular}
\caption{The ablation study of the amending generation on MultiWOZ 2.2 with joint goal accuracy. NS means the negative sampling mentioned in Section~\ref{sec:amending_generation}, as well as NS$^+$ performs a heuristic negative sampling.}\label{tab:ablation-amend}
\end{table}

\paragraph{Visualization} Figure~\ref{fig:attention} shows the attention visualization of the basic generation and the amending generation on an example of MultiWOZ 2.2. In the basic generation, the value of slot \textit{restaurant-area} is copied from previous dialogue state and the value of slot \textit{restaurant-people} is predicted from user utterance accurately. However, for the \textit{restaurant-name}, although the model pays attention to the corresponding tokens in the dialogue of the current turn, it generates a wrong slot value. In the amending generation, the \textit{restaurant-name} and its value attend to both the corresponding tokens in the user utterance and slot value in the primitive dialogue state with high weight, which indicates that the amending generation can utilize both the dialogue of the current turn and the primitive dialogue state for reference to correct the mistakes in the primitive dialogue state.

\begin{figure}
	\centering
	\includegraphics[width=0.49\textwidth]{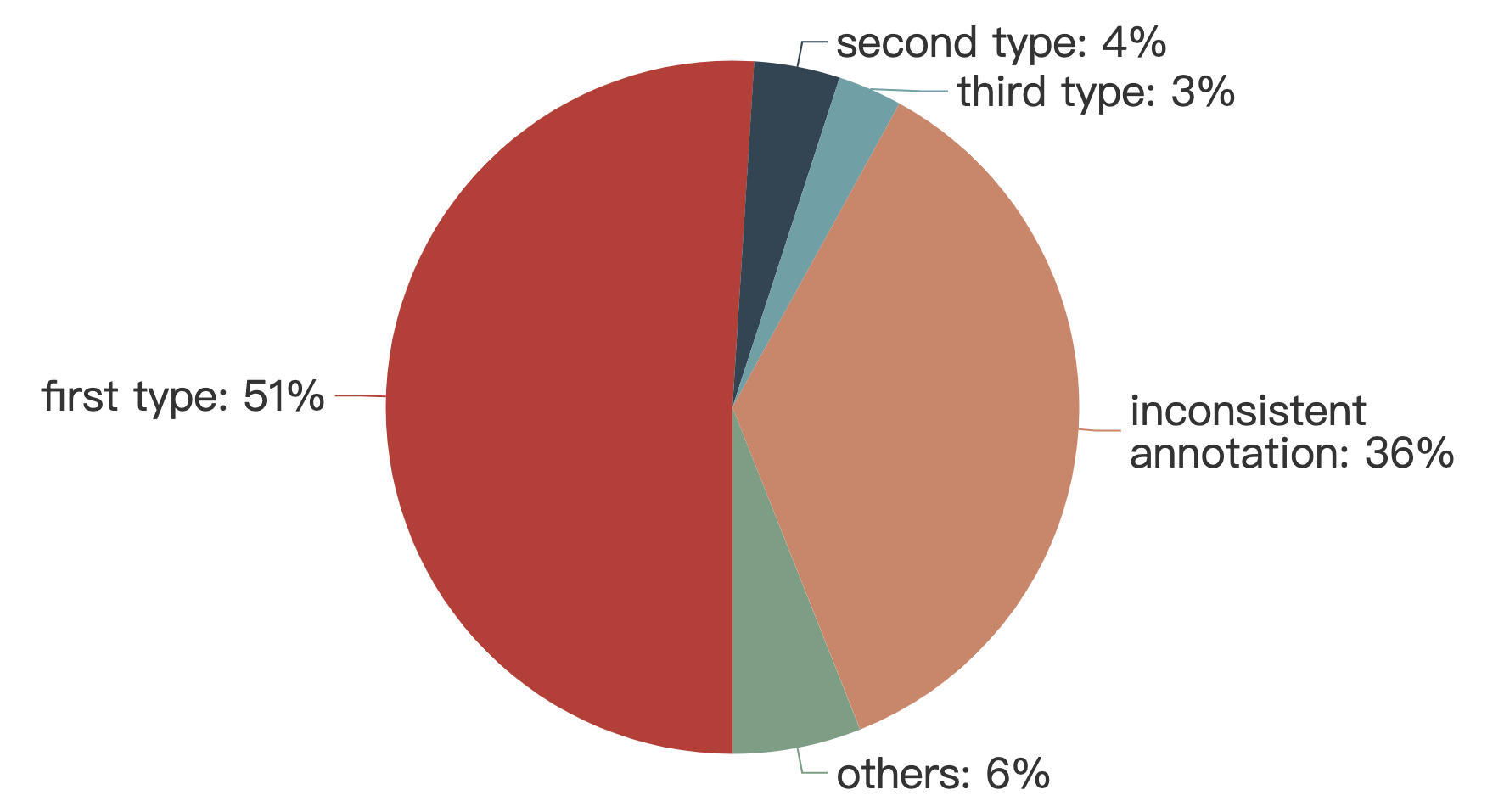}
	\caption{Error type statistics of the basic generation. first type: slot value not updated; second type: common mistakes of generation; third type: confusion between correlated slots.}
	\label{fig:error_type}
\end{figure}

\paragraph{Error Analysis} In our analysis, we found three frequent errors in a one-pass generation model of DST. (1) The slot value is not updated, as the model fails to obtain the key information in the dialogue, an example can be seen in Figure~\ref{fig:example}. (2) Some common mistakes occur in generation model. For example, \textit{cambridge lodge} is predicted to \textit{cambridge lodge lodge}, and \textit{camboats} is predicted to \textit{cambots}. (3) There is confusion between correlated slots, such as \textit{leave at} and \textit{arrive by}, \textit{departure} and \textit{destination}. As reported in Figure~\ref{fig:error_type}, we make the statistics of error types by random sampling on MultiWOZ 2.2, which indicates that 51\% of the error comes from the first type, 4\% comes from the second type, 3\% comes from the third type, 36\% is due to the inconsistent annotation, and 6\% is due to others. 
Furthermore, we show that the proposed amendable generation can greatly correct these errors and improve the overall performance of DST with concrete examples of the amending generation are shown in Table~\ref{tab:case-study} of Appendix~\ref{appendix:case_study}. 

\subsection{Effect of Previous Dialogue State}
As shown in Table~\ref{tab:structures}, we compare three types of input sequence formations: only dialogue history, dialogue history and previous dialogue state, current turn dialogue and previous dialogue state. We find that using the previous dialogue state (i.e. dialogue state memory) performs better than using only dialogue history, which confirms that the previous dialogue state can be served as a compact representation of the dialogue history and utilizing the previous dialogue state is more effective than performing DST from scratch. The results also show that using the current turn dialogue instead of the whole dialogue history barely hurts the performance, yet it greatly improves the efficiency, which justifies our use of the current turn dialogue and previous dialogue state as the input sequence.

\begin{table}
\centering
\begin{tabular}{lc}
\hline
\textbf{Structure} & \textbf{Joint Acc} \\
\hline
\hline
$(B_t|D)$ $^\dagger$ & 54.02 \\
$(B_t|D,B_{t-1})$ & \textbf{56.27} \\
$(B_t|D_t,B_{t-1})$ & 56.20 \\
\hline
\end{tabular}
\caption{Joint goal accuracy of different structures of generation model on MultiWOZ 2.2. The structures with dialogue state memory perform better than the rest which uses only dialogue context. $^\dagger$: we reuse the result of SimpleTOD in Table~\ref{tab:jga-baselines}.}\label{tab:structures}
\end{table}

\begin{table}
\centering
\small
\begin{tabular}{lc}
\hline
\textbf{Model} & \textbf{Joint Acc} \\
\hline
\hline
SimpleTOD & 54.02 \\
Seq2Seq-DU & 54.40 \\
\hline
GPT-2 w/ basic generation & 55.24 \\
GPT-2 w/ amending generation & 55.85 \\
GPT-2 w/ amending generation + NS & 56.09 \\
\hline
PLATO-2 w/ basic generation & 56.20 \\
PLATO-2 w/ amending generation & 57.07 \\
PLATO-2 w/ amending generation + NS & \textbf{57.26} \\
\hline
\end{tabular}
\caption{The ablation study of the pre-trained model on MultiWOZ 2.2 with joint goal accuracy.}\label{tab:ablation-pretrain}
\end{table}

\subsection{Effect of Pre-training}
In prior works, GPT-2 is often the pre-trained backbone in generation model~\citep{NEURIPS2020_e9462095,ham-etal-2020-end,yang2021ubar}. We also analyze the ability of GPT-2 on MultiWOZ 2.2 (see Table~\ref{tab:ablation-pretrain}), where only token embedding and position embedding are added as input embedding. Results show that our approach initialized with GPT-2 surpasses the SimpleTOD (a generation model initialized by GPT-2) and Seq2Seq-DU (the prior state-of-the-art), and obtains similar results with the PLATO-2 initialization. This indicates our approach's ability to universally improve the performances of other pretrained models.

\subsection{Effect of Special Tokens}
Special tokens are important for identifying different input components~\citep{NEURIPS2020_e9462095}. In our experiments, similar to prior works, we use the utterance special token and dialogue state special token to differentiate each part. Besides, to boost the extraction of entity names on MultiWOZ 2.2, we add special tokens \texttt{<name/>} and \texttt{</name>} around the candidate entity names. Table~\ref{tab:ablation-token} shows the joint goal accuracy reduction caused by the absence of various special tokens, which further confirms the neccesity of using special tokens.

\begin{table}
\centering
\begin{tabular}{ll}
\hline
\textbf{Model} & \textbf{Joint Acc} \\
\hline
\hline
basic generation & \textbf{56.20} \\
\hline
\quad - name special token & 55.34 (-0.86) \\
\quad - utterance special token & 55.70 (-0.50) \\
\quad - DS special token & 54.50 (-1.70) \\
\quad - above three & 54.41 (-1.79) \\
\hline
\end{tabular}
\caption{The ablation study of the absence of different special tokens on MultiWOZ 2.2 with joint goal accuracy.}\label{tab:ablation-token}
\end{table}

\section{Related Work}
Traditional methods regard DST as a classification task, in which an encoder is employed for obtaining a representation of utterances and a classifier is utilized for predicting a slot value from a pre-defined ontology~\citep{mrksic-etal-2017-neural,zhong-etal-2018-global,ramadan-etal-2018-large}. However, it is difficult to obtain a full ontology in a real scenario. Some open-vocabulary DST models extract or generate the dialogue state from the dialogue history at each turn~\citep{Chao2019,NEURIPS2020_e9462095,ham-etal-2020-end,heck-etal-2020-trippy}. These methods predict the dialogue state from scratch at each turn, which undoubtedly puts a great burden on the model.

To address this problem, some recent methods consider to utilize the previous dialogue state and recent dialogue history as input information. Specifically, the DST is separated into two components, where a component named State Operation Prediction is used to encode utterance and previous dialogue state, as well as predict the operation of state at each turn, and the other component named Value Generation predicts the value for each slot~\citep{kim-etal-2020-efficient,zeng2020multi}. 
\citet{kim-etal-2020-efficient} encodes the dialogue history of the last two turns and the previous dialogue state by BERT and predicts four kinds of state operations: \textit{carryover}, \textit{delete}, \textit{dontcare} and \textit{update}. For \textit{update} operation, a GRU-based value generation is used to decode the slot value.
\citet{zeng2020multi} reuses a fraction of the hidden states of the encoder in the decoder to build a flat model for effective parameter updating.
Besides, some methods treat dialogue state tracking as a causal language model by using the dialogue of the current turn and previous dialogue state as input sequence~\citep{lin-etal-2020-mintl,yang2021ubar}. \citet{lin-etal-2020-mintl} utilizes an encoder-decoder framework to generate dialogue state and system response sequentially, where minimal slot value pairs are generated for efficiently tracking. \citet{yang2021ubar} models task-oriented dialogs on a dialog session level, which generates dialogue state, dialogue action and system response sequentially based on the whole previous dialogue context, including the generated dialogue states and dialogue actions. 
Moreover, some schema-guided DST methods leverage schema descriptions to deal with unseen schemas with new domains and slots~\citep{zhu-etal-2020-efficient,feng-etal-2021-sequence,noroozi2020fast}.


\section{Conclusion}

In this paper, we propose a novel amendable generative approach for dialogue state tracking, which learns implicit state operation prediction and value generation jointly in a single model to reduce the error propagation. Meanwhile, our model offers an opportunity to amend the primitive dialogue state in the amending generation. Experimental results show that our model outperforms previous works on both MultiWOZ 2.2 and WOZ 2.0 datasets, achieving state-of-the-art performance. The ablation study and attention visualization demonstrate that the proposed amending generation is significantly effective. Moreover, we analyze the types of mistakes that can be resolved by the amending generation and provide some examples to illustrate them in the Appendix. In the future, we will try to integrate schema descriptions into our architecture and explore a generative approach to support end-to-end task-oriented dialogue system.

\bibliography{anthology,custom}
\bibliographystyle{acl_natbib}

\clearpage
\appendix

\section{Error Propagation Analysis}

Our model learns simultaneously the implicit state operation prediction and value generation by a single generation model with dialogue state memory. Indeed, such joint learning can boost the performance of the \textit{update} gate. Because the traditional method needs to predict the state operation firstly, and then use another component to predict the corresponding value, which leads to error propagation. The \textit{update} gate accuracy and the value generation F1 under the \textit{update} gate are given in Table~\ref{tab:gate}.\footnote{Note that for our model, the state operation can be concluded from the difference between last and current dialogue states.} Compared with the SOM-DST, our AG-DST attains better results of both gate accuracy and value F1. This implies that the coupling structure we used is beneficial to learning \textit{update} gate and corresponding value prediction.

\begin{figure*}[bp]
	\centering
	\subfloat[\textit{update} gate\label{fig:attention_update}]{\includegraphics[width=0.449\textwidth]{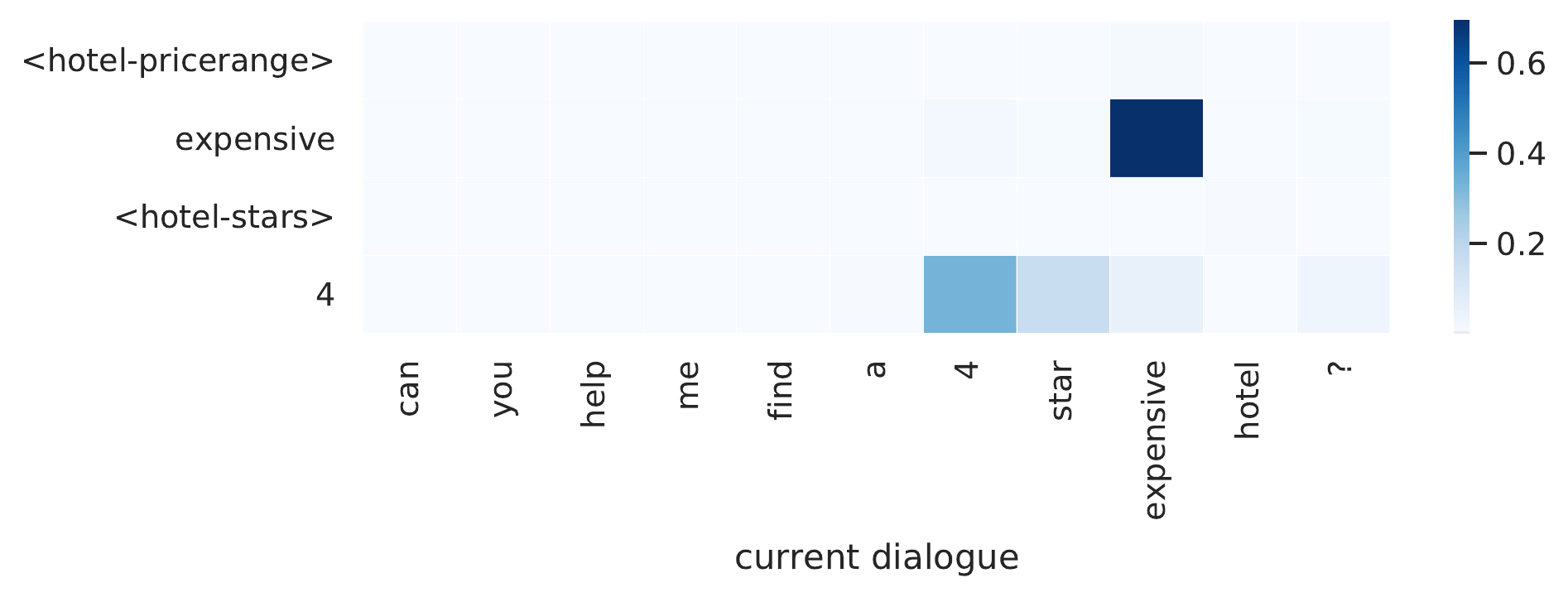}}\hspace{-0.5em}
	\subfloat[\textit{dontcare} gate\label{fig:attention_dontcare}]{\includegraphics[width=0.549\textwidth]{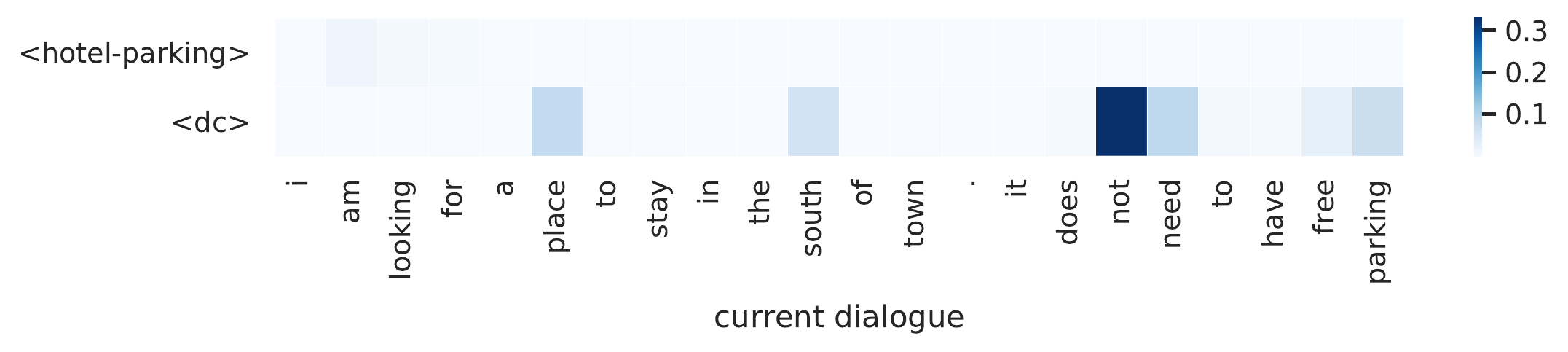}}\\\vspace{-0.5em}
	\subfloat[\textit{carryover} gate\label{fig:attention_carryover}]{\includegraphics[width=0.65\textwidth]{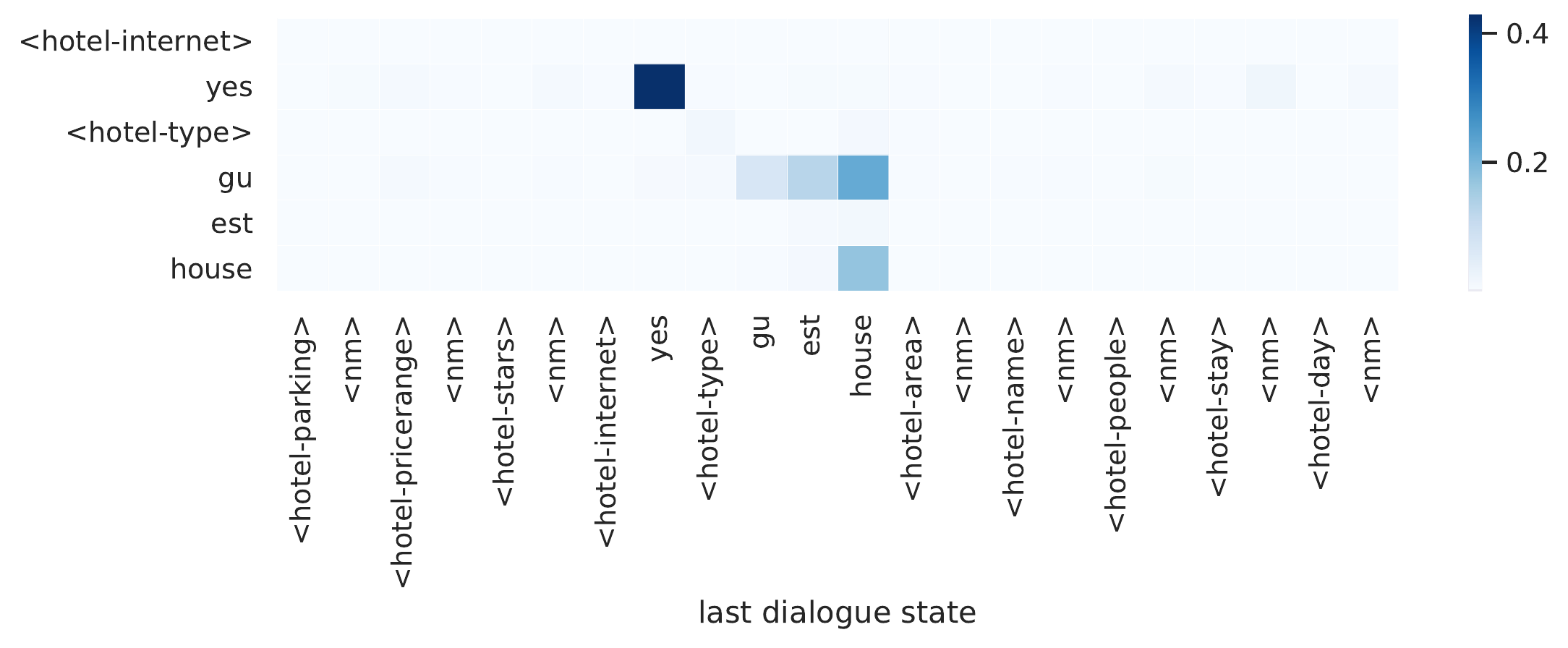}}
	\caption{The attention visualization of state operation on MultiWOZ 2.2.}
	\label{fig:attention_gate}
\end{figure*}

\section{Case Study}\label{appendix:case_study}

Table~\ref{tab:case-study} indicates the the amendable ability of the AG-DST. On the test set of MultiWOZ 2.2, the total number of improved examples for the three error types is 313. AG-DST was able to amend the slot value during the amending generation, when the basic generation decoded nothing or a wrong value.

\section{Attention Visualization of State Operation Prediction}

\begin{figure*}
	\centering
	\subfloat[\textit{delete} gate\label{fig:attention_delete}]{\includegraphics[width=0.22\textwidth]{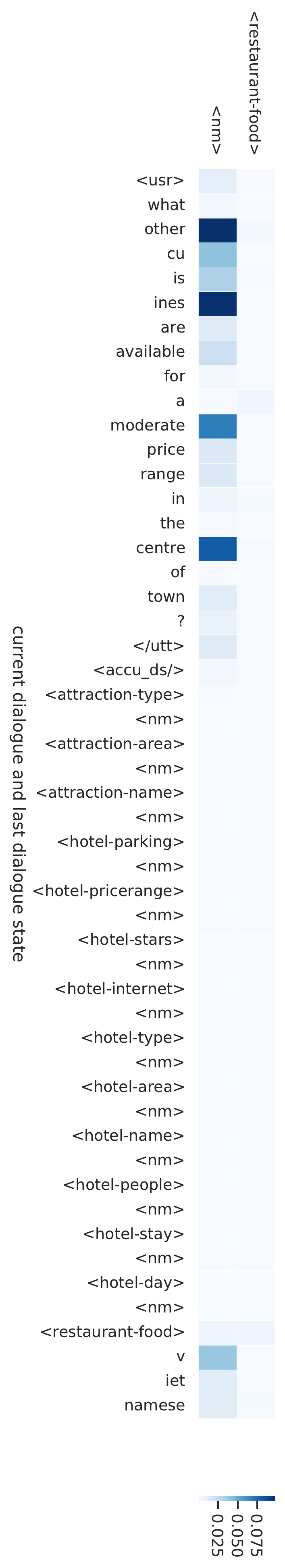}}\hspace{4em}
	\subfloat[\textit{coreference} phenomenon\label{fig:attention_coreference}]{\includegraphics[width=0.38\textwidth]{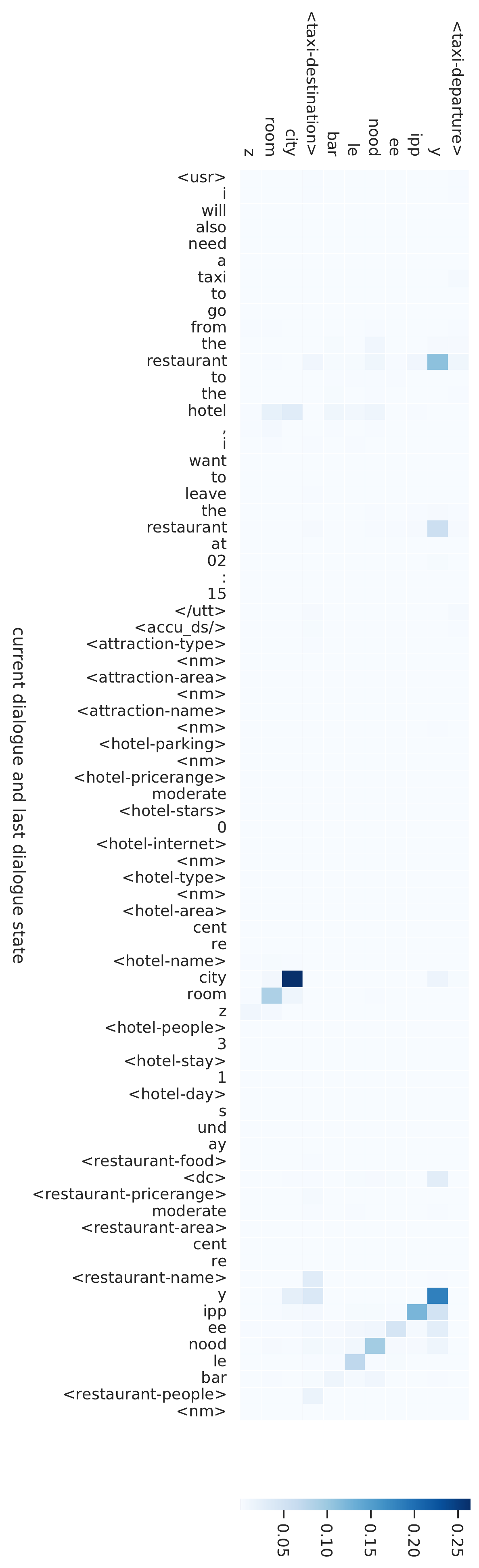}}
	\caption{The attention visualization of \textit{delete} gate and \textit{coreference} phenomenon on MultiWOZ 2.2.}
	\label{fig:attention_gate2}
\end{figure*}

Figure~\ref{fig:attention_gate} shows the attention visualization of different state operations on test set of MultiWOZ 2.2. From the \textit{update} gate and \textit{dontcare} gate attention visualizations in Figure~\ref{fig:attention_update} and~\ref{fig:attention_dontcare}, we can easily find that the slot values attend to the corresponding tokens in the utterance with a highest weight. Figure~\ref{fig:attention_carryover} indicates that an implicit copy mechanism is used for the \textit{carryover} gate. Figure~\ref{fig:attention_gate2} shows the attention visualization of \textit{delete} gate and \textit{coreference} phenomenon. For \textit{delete} gate in Figure~\ref{fig:attention_delete}, it is easily seen that the model can grasp the information in the user utterance to generate appropriate \texttt{<nm>} slot value. As shown in Figure~\ref{fig:attention_coreference}, when the user only said ``from the restaurant to the hotel'', the model can extract the corresponding restaurant and hotel names from DS memory. This indicates that our model is able to handle the \textit{coreference} operation. Most examples of this \textit{coreference} phenomenon appear in \textit{taxi} domain.

\begin{table}
\centering
\begin{tabular}{lcc}
\hline
\textbf{Model} & \textbf{Acc} & \textbf{F1} \\
\hline
\hline
SOM-DST & 70.50 & 78.58 \\
AG-DST & \textbf{72.77} & \textbf{79.84} \\
\hline
\end{tabular}
\caption{The comparison of \textit{update} gate accuracy and corresponding value generation F1 between SOM-DST~\citep{kim-etal-2020-efficient} and our basic generation on MultiWOZ 2.2.}\label{tab:gate}
\end{table}

\begin{table*}
\centering
\fontsize{6.5pt}{0.6\baselineskip}\selectfont
\begin{tabular}{p{6cm}|p{2.5cm}|p{3cm}|p{3cm}}
\toprule
\textbf{Current Turn Dialogue} & \textbf{Previous Dialogue State} $\hat{B}_{t-1}$ & \textbf{Primitive Dialogue State} $\tilde{B}_t$ & \textbf{Amended Dialogue State} $\hat{B}_t$ \\
\midrule
$R_t$: no , i am sorry . i am searching for a 4 star hotel in the centre for 2 nights on saturday . is that correct ? & hotel-stay-2 & hotel-stay-2 & hotel-stay-3 \\
$U_t$: i need the hotel for \textcolor{my_blue}{\textit{3 nights}} starting on saturday please . & & & \\
\midrule
$R_t$: ok , and what day and time would you like that reservation ? & restaurant-people-2 & restaurant-people-2 & restaurant-people-1 \\
$U_t$: i would like to make a reservation for saturday at 11:45 . and there has been \textcolor{my_blue}{\textit{a change in plans , i will be dining alone}} . & & & \\
\midrule
$R_t$: hello . & restaurant-food-\texttt{<nm>} & restaurant-food-indian & restaurant-food-south indian \\
$U_t$: i am looking for info on expensive \textcolor{my_blue}{\textit{south indian}} restaurant -s in cambridge . & & & \\
\midrule
$R_t$: ok , so you would like a taxi from the restaurant to the park ? could you please let me know your desired departure and arrival times ? & taxi-departure-byard art & taxi-departure-byard art & taxi-departure-wandlebury country park \\
$U_t$: i am sorry , i would like a taxi from \textcolor{my_blue}{\textit{wandlebury country park}} to taj tandoori . i would like the taxi to pick me up at 10:15 . & & & \\
\midrule
$R_t$: yes , \texttt{<name/>} \textcolor{my_blue}{\textit{wandlebury country park}} \texttt{</name>} is in the south . in order to help you book a taxi between the park and your hotel , i need to know what hotel you are at . & taxi-destination-kohinoor & taxi-destination-kohinoor & taxi-destination-wandlebury country park \\
$U_t$: i want a taxi from the restaurant that i am at & & & \\
\midrule
$R_t$: i am sorry , i am experiencing a system error . could you please restate your request ? & attraction-type-swimmingpool & attraction-type-entertainment & attraction-type-swimmingpool \\
$U_t$: i want a place to go in the south , \textcolor{my_blue}{\textit{a swimmingpool . or another type of entertainment}} , if there is no pool ? & & & \\
\midrule
$R_t$: \texttt{<name/>} caffe uno \texttt{</name>} is a very nice , expensive italian restaurant in the center of town . would you like a table there ? & restaurant-name-clowns & restaurant-name-clowns & restaurant-name-\texttt{<nm>} \\
$U_t$: actually , \textcolor{my_blue}{\textit{i change my mind}} . i think i want to stick with british food after all . can you suggest any 1 thats in the centre of town ? & & & \\
\midrule
$R_t$: \texttt{<name/>} primavera \texttt{</name>} is a museum in the center of town . their address is 10 king s parade . what else can i help with ? & restaurant-time-17:45 & restaurant-time-17:45 & restaurant-time-16:15 \\
$U_t$: oh , i made a mistake . i really need that table \textcolor{my_blue}{\textit{for 16:15 , not 17:45}} . can you change it , do you thing ? & & & \\
\midrule
$R_t$: there are 5 museums in the west . i recommend \textcolor{my_blue}{\textit{kettle's yard}} . would you like the address and phone ? & attraction-name-cambridge punter & attraction-name-cambridge punter & attraction-name-kettles yard \\
$U_t$: yes , i would love the address . thank you so much ! & & & \\
\midrule
$R_t$: you would like a second hotel , for which night and location ? & taxi-departure-\texttt{<nm>} & taxi-departure-\texttt{<nm>} & taxi-departure-alexander bed and breakfast \\
$U_t$: i am sorry , i do not want a second hotel . i need a taxi between \textcolor{my_blue}{\textit{the bed and breakfast}} and the restaurant that arrives to the restaurant by the booked time . & & & \\
\midrule
$R_t$: to clarify you wanted me to book a table on friday ? & train-day-\texttt{<nm>} & train-day-\texttt{<nm>} & train-day-tuesday \\
$U_t$: i am sorry . my train was supposed to be for \textcolor{my_blue}{\textit{tuesday}} as well as my restaurant reservation . & & & \\
\midrule
$R_t$: there are 5 choices . to narrow down you need to choose the side of town you prefer & hotel-area-\texttt{<nm>} & hotel-area-\texttt{<nm>} & hotel-area-\texttt{<dc>} \\
$U_t$: \textcolor{my_blue}{\textit{area does not matter}} , but it should be a guest house with a 4 star rating . & & & \\
\midrule
$R_t$: the entrance fee is unknown . can i help you with anything else today ? & taxi-destination-\texttt{<nm>} & taxi-destination-tajione restaurant and coffee bar & taxi-destination-stazione restaurant and coffee bar \\
$U_t$: i would also like a taxi to \textcolor{my_blue}{\textit{arrive at the restaurant}} on time . can i have the taxis phone number and vehicle type ? & & & \\
\midrule
$R_t$: okay ! they are at king's parade , cb21rl . their phone number is 01223338300 . what time would you like to depart in the taxi ? & taxi-destination-\texttt{<nm>} & taxi-destination-saint catharine's college & taxi-destination-saint catharines college \\
$U_t$: i need a taxi to \textcolor{my_blue}{\textit{commute between the 2 place -s . i need to leave the restaurant}} by 18:15 and need the contact \# and car type & & & \\
\midrule
$R_t$: i have got you a reservation for 6 at \textcolor{my_blue}{\textit{hobson's house}} for 2 nights . your reference number is 4wngilmf . & hotel-name-\texttt{<nm>} & hotel-name-hobson's house & hotel-name-hobsons house \\
$U_t$: thank you so much ! that should be all i need . & & & \\
\midrule
$R_t$: is there anything i can do for you ? & taxi-destination-\texttt{<nm>} & taxi-destination-caffe jello gallery & taxi-destination-cafe jello gallery \\
$U_t$: i would like a taxi to the \textcolor{my_blue}{\textit{cafe jello gallery}} please . & & & \\
\midrule
$R_t$: \texttt{<name/>} \textcolor{my_blue}{\textit{camboats}} \texttt{</name>} is located at the plough , green end , fen ditton , in the east side of town . & taxi-destination-\texttt{<nm>} & taxi-destination-cambots & taxi-destination-camboats \\
$U_t$: ok ! thank you ! can you get me a taxi to cambots ? & & & \\
\midrule
$R_t$: the postcode is cb28rj for \texttt{<name/>} \textcolor{my_blue}{\textit{bridge guest house}} \texttt{</name>} . what time do you want to be \textcolor{my_blue}{\textit{picked up at the guest house}} or to arrive at eraina ? & taxi-destination-eraina & taxi-destination-eraina & taxi-destination-bridge guest house \\
$U_t$: okay i need a taxi too . & & & \\
\bottomrule
\end{tabular}
\caption{Examples of the amending generation. The key information in the dialogue is shown in blue.}\label{tab:case-study}
\end{table*}

\end{document}